\documentclass[conference]{IEEEtran}
\IEEEoverridecommandlockouts
\usepackage{cite}
\usepackage{amsmath,amssymb,amsfonts}
\usepackage{algorithmic}
\usepackage{graphicx}
\usepackage{textcomp}
\usepackage{xcolor}
\usepackage[utf8]{inputenc}
\usepackage{dblfloatfix}
\usepackage{url}
\usepackage{makecell}
\usepackage{subcaption}
\usepackage{tabularx}
\newcolumntype{L}{>{\centering\arraybackslash}X}
\usepackage{tikz}
\usepackage{pgfplots}
\pgfplotsset{compat=1.13}
\def\BibTeX{{\rm B\kern-.05em{\sc i\kern-.025em b}\kern-.08em
    T\kern-.1667em\lower.7ex\hbox{E}\kern-.125emX}}
\begin{document}

\title{Solving Raven's Progressive Matrices with Multi-Layer Relation Networks}

\author{\IEEEauthorblockN{Marius Jahrens}
\IEEEauthorblockA{\textit{Institute for Neuro- and Bioinformatics} \\
\textit{University of Lübeck}\\
Lübeck, Germany \\
jahrens@inb.uni-luebeck.de}
\and
\IEEEauthorblockN{ Thomas Martinetz}
\IEEEauthorblockA{\textit{Institute for Neuro- and Bioinformatics} \\
\textit{University of Lübeck}\\
Lübeck, Germany \\
martinetz@inb.uni-luebeck.de}
}

\maketitle

\begin{abstract}
 Raven's Progressive Matrices are a benchmark originally designed to test the cognitive abilities of humans. It has recently been adapted to test relational reasoning in machine learning systems. For this purpose the so-called Procedurally Generated Matrices dataset was set up, which is so far one of the most difficult relational reasoning benchmarks. Here we show that deep neural networks are capable of solving this benchmark, reaching an accuracy of 98.0 percent over the previous state-of-the-art of 62.6 percent by combining Wild Relation Networks with Multi-Layer Relation Networks and introducing Magnitude Encoding, an encoding scheme designed for late fusion architectures.
\end{abstract}

\begin{IEEEkeywords}
Raven's Progressive Matrices, Procedurally Generated Matrices (PGM), Wild Relation Networks, Multi-Layer Relation Networks
\end{IEEEkeywords}

\section{Introduction and Previous Work}
Intelligent behaviour requires the ability to reason about relations. Several relational reasoning benchmarks for machine learning were proposed but have been cleared in the meantime by neural network based approaches with high accuracies in excess of 95\%, e.g. the visual question answering benchmark CLEVR \cite{johnson16} by \cite{santoro17} and the text-based question answering benchmark bAbI \cite{weston15} by \cite{jahrens19}. Both benchmarks are challenging also for humans. So in an attempt to pose a new, harder challenge, the Procedurally Generated Matrices (PGM) dataset \cite{barrett18a} was proposed. It is based on Raven's Progressive Matrices \cite{raven38}, which were originally designed to test cognitive abilities of humans independent from their level of education. It is considered to measure abstract reasoning and so-called fluid intelligence. The authors of the PGM dataset also presented a couple of neural network models with state-of-the-art architectures to solve the new benchmark. However, they achieved only a maximum of 62.6\% accuracy under the most general conditions, hence leaving room for improvements.

Unpublished work by \cite{steenbrugge18} improve on the results of the PGM authors by generating disentangled feature representations via a Variational Autoencoder. With this representation the test accuracy increases to 64.2\%. Their main contribution, however, lies in showing that disentangled representations cause less of a performance drop when a subset of the relations is withheld from the training domain. An Attention Relation Network (ARNe) combining Relation Networks with attention mechanisms was proposed by \cite{hahne19}. In contrast to this work, their training uses meta labels as additional training signals, allowing it to achieve 88.2\% test accuracy, but making the results not directly comparable. 

The Relation Network (RN) module introduced in \cite{santoro17} has proven to be a powerful component for solving these kinds of problems \cite{santoro17}\cite{jahrens19}\cite{barrett18a}. So, rather than exploring alternative architectures, in this work we will demonstrate that the existing RN based models can be extended to solve the PGM dataset with 98.0\% test accuracy, given the right training method\footnote{As supplementary material, the source code to reproduce our results is available at: \url{http://webmail.inb.uni-luebeck.de/exchange-supplement/PGM_MLRN_supplementary.zip}.}.

\section{Procedurally Generated Matrices}
\begin{figure*}
\centering
\begin{subfigure}{0.45\textwidth}
    \centering
    \includegraphics[width=0.54\linewidth]{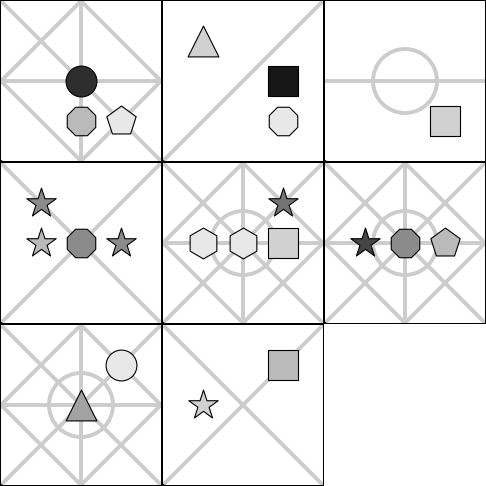}
    \caption{Context}
    \label{fig:samplecontext}
\end{subfigure}%
\begin{subfigure}{0.45\textwidth}
    \centering
    \includegraphics[width=0.72\linewidth]{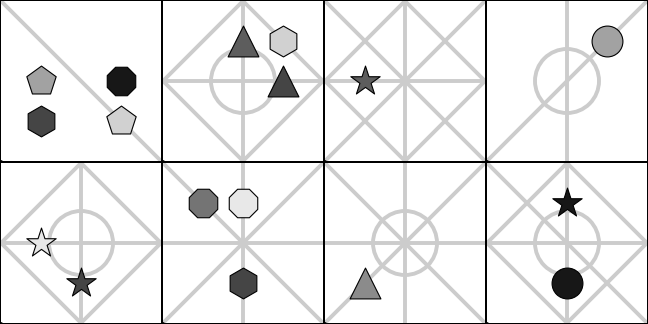}
    \caption{Options}
    \label{fig:sampleoptions}
\end{subfigure}
\caption{A sample from the PGM dataset. In each row of the context the third image has a shape only at the positions where both the first and the second image in the row have shapes. The correct answer is the top right option. The shape types, colors and the lines in the background are distractors. This sample is an example for the structure triple \mbox{($object\_type=shape$, $attribute\_type=position$, $relation\_type=\text{AND}$)}.}
\label{fig:datasample}
\end{figure*}

Like the original Raven's Progressive Matrices, PGM consists of image sequences that are to be completed logically.
Every sample consists of eight images for context and eight answer images to choose from, as shown in Fig. \ref{fig:datasample}. 
The context forms a 3x3 grid of images where the last image is missing and needs to be filled in by choosing one from the eight answer options. 
The images in the grid are related row-wise or column-wise through one or more triples of the form $(object\_type, attribute\_type, relation\_type)$ specifying the kind of relationship. 
\begin{itemize}
    \item $object\_type \in \{line, shape\}$, whether the subject is the lines in the background or the symbols in the foreground
    \item $attribute\_type \in \{color, position, type, number, size\}$, the attribute the relation applies to
    \item \begin{flushleft}$relation\_type \in \{\text{AND}, \text{OR}, \text{XOR}, progression, \allowbreak consistent\ union\}$, the relation type\end{flushleft}
\end{itemize}

In the paper detailing the PGM dataset \cite{barrett18a}, three different benchmarking modalities are suggested: 

\begin{enumerate}
\item The image sequences may or may not include distractors, i.e. image elements or properties which do not hold any information about the underlying relation. In this paper the full dataset with distractors is used. 

\item The dataset offers different generalization regimes with entire classes of problems withheld from the training data.
We focus on the neutral regime with the training data being representative of the test data.

\item Training labels may or may not include details about the underlying relation, serving as additional hints/training signals. 
No such information was used in this paper.
\end{enumerate}

The dataset has 1.2 million training samples, a validation set of size 20k, and 200k test samples.

\section{Wild Relation Network}
\begin{figure*}
\centering
\includegraphics[width=\linewidth]{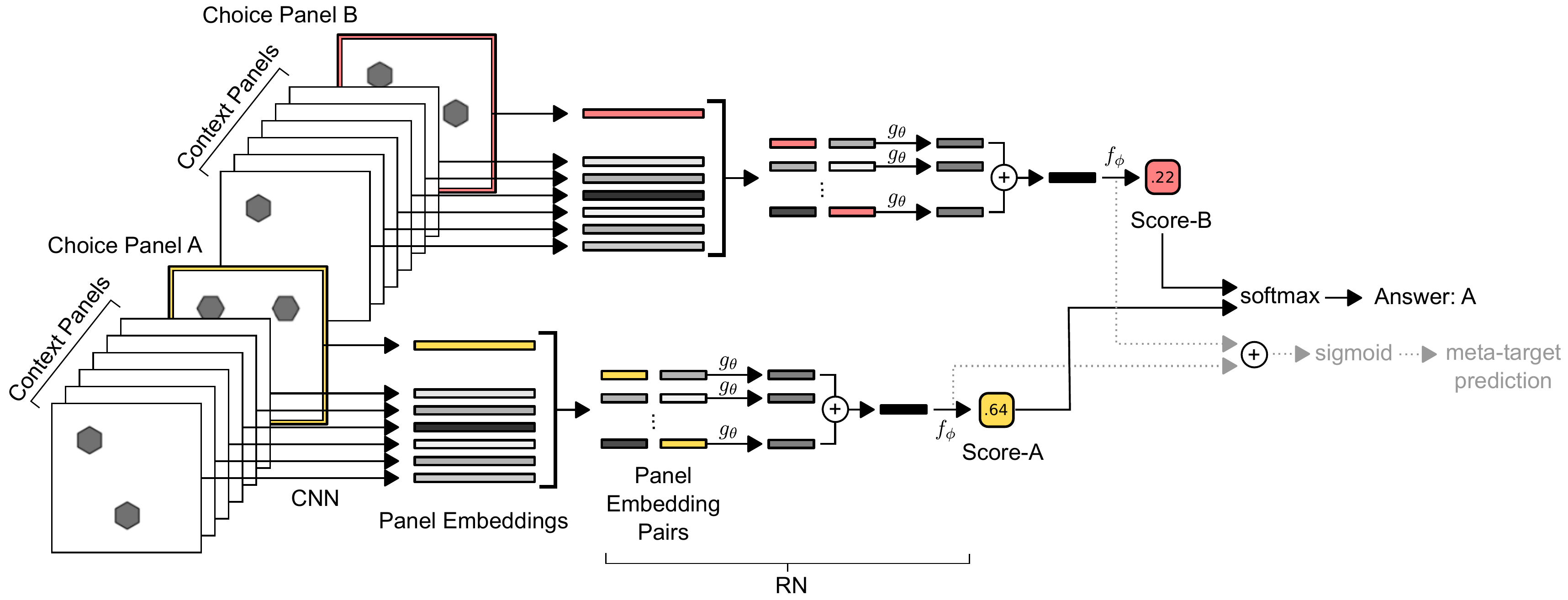}
\caption{The Wild Relation Network for solving the PGM benchmark. A CNN module forms panel embeddings for each panel image from context and choice. Panel embedding pairs are the input for a Relation Network (RN) module. Each choice option is scored independently, and the option with the highest score is chosen as answer (figure taken from  \cite{barrett18a} with permission from the authors).}
\label{fig:wren}
\end{figure*}

The Wild Relation Network (WReN) was proposed by \cite{barrett18a} as a neural network based approach for solving the PGM benchmark. Each of the eight choice options is scored independently, as shown in Fig. \ref{fig:wren}. Each context and choice panel image is fed into a Convolutional Neural Network (CNN) and represented by an embedding vector, which also includes a 9-dimensional one-hot vector to encode the image position within the 3x3 grid. For every score, embeddings of the eight context images as well as an embedding of one of the answer options are used to form pairwise inputs for a Relation Network module, as introduced in \cite{santoro17}. The whole neural network consisting of CNN and RN is trained end-to-end. 

The authors have shown that the RN module at the core of the model makes a crucial contribution to its performance. 
However, the single pairwise evaluation has been shown to be a limiting factor in the past \cite{jahrens19}. We will see that this is also one reason for the suboptimal results achieved in \cite{barrett18a}. In the following we will introduce several measures for significant improvements.

\section{Measures for improvement}
\subsection{Wild Multi-Layer Relation Networks}

\begin{figure*}
\centering
\includegraphics[width=0.9\linewidth]{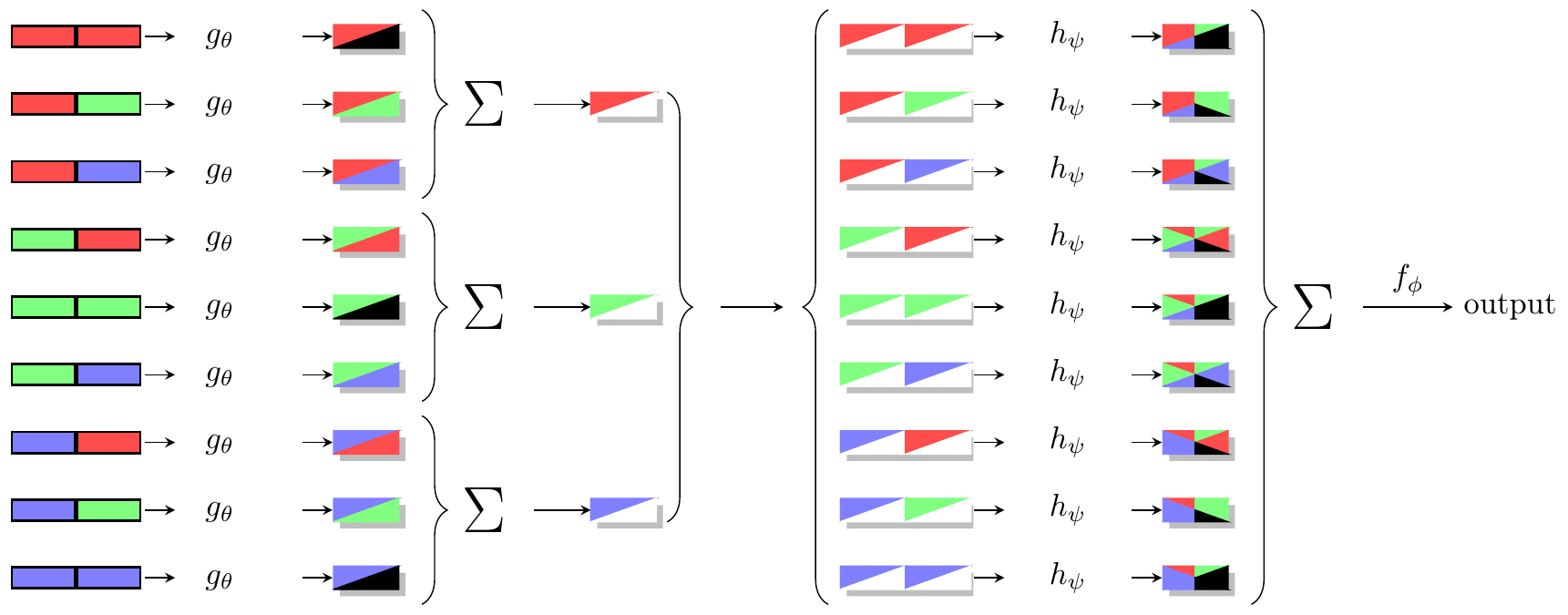}
\caption{A Multi-Layer Relation Network with two layers: On the left the relations for pairs of image embeddings (red, green, blue) are evaluated and all relations for the same base image are combined to form new intermediate representations (red/white, green/white, blue/white). The intermediate representations are then used as input for the next relational layer on the right.}
\label{fig:dlrn}
\end{figure*}

To be able to learn more complex relations, in \cite{jahrens19} the RN was extended to a Multi-Layer Relation Network (MLRN). This was shown to be crucial for solving some of the tasks in the  bAbI benchmark \cite{weston15}, \cite{jahrens19}. We expect this extension also to be crucial in the PGM benchmark. While the vanilla RN reduces all pairwise results to a single vector, the MLRN only combines results relating to the same image into the same vector. 
That way a new representation of each image is generated, enriched with information about how it relates to the other images in the grid. 
These new representations are then used as inputs for the next relational layer, as shown in Fig. \ref{fig:dlrn}. The Wild Multi-Layer Relation Network is then based on the MLRN analogous to the single-layer WReN based on vanilla RN.

\subsection{Regularization}
Originally, the WReN model in \cite{barrett18a} was trained with dropout. However, RNs and more generally MLRNs have previously been shown to work well with L2 regularization on the weights for other problems \cite{santoro17}\cite{jahrens19}. Therefore, we apply L2 regularization as well and compare it with dropout. 

\subsection{Magnitude Encoding}

As we will see later in the experiments, samples incorporating the $color$ attribute have a much lower accuracy than all other attributes, and additional relational layers do not help to improve it. Since the features of multiple images are only fused from the relation layers onward, we conjecture that the color information is not well preserved in the image embeddings. 
The relational layers using the image embeddings seem to lack information to determine the change in colors across images. To test this hypothesis, we encoded the intensities of the grayscale images in such a way, that different intensity levels are reflected in different input features, rather than a single scalar per pixel. We will call this encoding scheme Magnitude Encoding (ME) and denote the dimensionality $d$ of the vector representation as ME$d$, e.g. ME20 for $d=20$.  

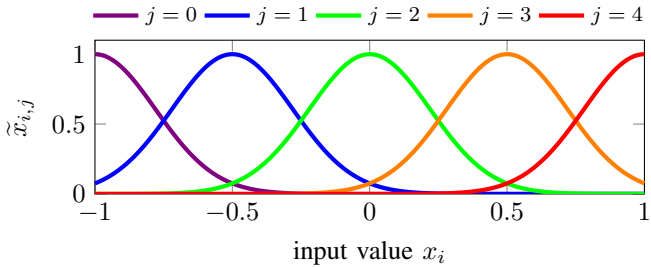
\begin{figure}
    \centering
    \begin{tikzpicture}
        \begin{axis}[
            xlabel = input value $x_i$,
            ylabel = $\widetilde{x}_{i,j}$,
            xmin = -1,
            xmax = 1,
            xtick = {-1, -0.5, 0, 0.5, 1},
            ymin = 0,
            legend style = {font =\footnotesize, at={(0.5,1.03)}, anchor=south, draw=none},
            legend columns = 5,
            height=0.2\textwidth,
            width=0.49\textwidth,
            samples=200,
            domain=-1:1,
            cycle list name=ColorCyclelist
        ]
        \pgfplotsforeachungrouped \j in {0,...,4} {
        \addplot+[] {exp(-pow(x-(2*(\j)/4-1),2)/(2*pow(0.22,2)))};
        \addlegendentryexpanded{$j=\j$}
        }
        \end{axis}
    \end{tikzpicture}
    \caption{Visualization of magnitude encoding with $d=5$ dimensions. The index j represents the vector components $\widetilde{x}_{i,j}$ for the vectorial representation $\widetilde{x}_i$ of scalar value $x_i$.}
    \label{fig:me}
\end{figure}

Assuming samples $x$ being in the domain $x \in [-1, 1]^{n}$, then every scalar input $x_i$ is encoded independently via the gaussian encoding:
\begin{equation}
    \label{eq:gauss}
    \widetilde{x}_{i,j}=exp\left(-\frac{(x_i - (2\frac{j}{d-1}-1))^2}{2\sigma^2}\right)
\end{equation}
The encoding yields a tensor $\widetilde{x} \in [0, 1]^{n\times d}$ holding the vectorial representations of the input scalars. The hyperparameter $\sigma$ can be chosen freely, however, while a smaller value gives a sharper signal, this also reduces the number of weights of neighboring values that can learn from the sample. A visual representation of the scheme is shown in \mbox{Fig. \ref{fig:me}}. 

The reasoning behind the gaussian encoding is that all vector entries contain contributions from the encoded value. This can be especially useful if at any point ME is to be used to encode features inside a network instead of the network inputs, so that all components can carry gradients. For this dataset encoding the intensities as 20-dimensional vectors has proven to work well. Using larger vector representations did not yield any change in performance over ME20, and ME10 performed worse than ME20, though still better than not using ME at all. In the listed results only ME20 is used.

A similar encoding can be achieved with ReLUs. Then instead of Gaussians, for the encoding triangles are constructed out of the ReLUs. Both schemes show equal performance. The reported results are based on the Gaussian ME. 

\pgfplotscreateplotcyclelist{ColorCyclelist}{%
  {color=violet, mark = none, ultra thick},
  {color=blue, mark = none, ultra thick},
  {color=green, mark = none, ultra thick},
  {color=orange, mark = none, ultra thick},
  {color=red, mark = none, ultra thick},
}

\subsection{Alternative Optimizer}

The WReN model in \cite{barrett18a} was trained with the adaptive learning rate optimizer Adam \cite{kingma14}. As we will see later, the training and validation loss curves tend to exhibit a considerable amount of noise with this architecture. When inspecting the activations in different relational layers, a difference in the activation levels of multiple orders of magnitude can be observed. Therefore, we evaluate the layer-wise adaptive moments optimizer LAMB \cite{you19} as an alternative to the Adam optimizer. The LAMB optimizer copes with highly varying  activation levels by normalizing the corresponding gradients by each layer's weights' norm.

When using the LAMB optimizer, an additional warmup period for the learning rate as well as additional activation penalty loss terms are required in order for weights to not approach infinity, especially when using mixed precision training. We will see that the training and validation loss curves are much smoother compared to the training using Adam.

\section{Experiments}

\subsection{Architecture and Training}
The images in the PGM dataset have a resolution of \mbox{160 $\times$ 160} pixels. 
However, downscaled versions with \mbox{80 $\times$ 80} pixels are used instead, since the lower resolution does not harm the models' performance but reduces the computational cost. 

For magnitude encoding $\sigma=0.28$ is chosen.
The CNN for generating image embeddings has four convolution layers, each generating 32 feature maps using 3 $\times$ 3 kernels and a stride of 2.
The CNN's output is transformed into a 247 dimensional vector by a single linear layer and concatenated with a 9-dimensional one-hot vector to encode the image position in the grid, yielding a 256 dimensional embedding vector per image.

The first relation layer uses a multi-layer perceptron (MLP) with $[512, 512, 512, 256]$ neurons in the respective layers, while the second and third relation layers both have MLPs with $[256, 256, 256]$ neurons. 
For the final MLP $f_\phi$ producing the score, an MLP with $[256, 256, 1]$ neurons is used.

The batch size is 512 and the learning rate when using the LAMB optimizer is 2e-3 with a weight decay factor of 2e-1 on all weights (not applied to biases). 
Gradients are clipped to 1e1, and the clipping inside LAMB is deactivated. The optimizer's trust ratio's denominator has an offset of \mbox{1e-6} to avoid division by zero.
Since mixed precision training was used, a warmup period of 8 epochs is applied, that is linearly scaling the learning rate up on every iteration.
Finally, an activation loss term is added for the activations in the inputs and outputs of $f_\phi$, the last MLP in the model, which helps avoiding weights approaching infinity, especially during mixed precision training. 
The activation loss uses the mean square of the activations and adds to the total loss with factor 2e-3.

Training the models takes about 240 epochs, which equals 27 hours for the 3-layer model on four RTX 2080 Ti graphics cards. 

\subsection{Results}

Table \ref{tab:finalres} presents the improvements by the different measures listed above compared to the original results by \cite{barrett18a} with WReN (first column in Table \ref{tab:finalres}). The second column shows the results with L2 regularization instead of dropout. Keeping everything the same as in \cite{barrett18a} except for the regularization, the performance increases already drastically. The total error is reduced by a factor of 2. Only samples based on the $color$, $shape$, or $size$ attributes and samples based on XOR relations still pose a problem. 

While the XOR relation is by far the most difficult relation type for the single layer architecture, Table \ref{tab:finalres} shows that introducing additional relational layers closes the gap. With two additional layers in the Relation Network mainly the XOR problem class wins (column 3). This reinforces the notion from previous observations \cite{jahrens19}, that deeper MLRNs promote learning of more complex relations.

The main additional improvement is achieved when using ME for the $color$ attribute. The residual error is reduced by a factor of 4 (from column 3 to column 4). This supports the conjecture that using vectorial representations of scalar inputs with architectures employing late fusion of inputs is crucial. It is worth noting that with the LAMB optimizer the single layer RN is able to learn samples with the $color$ attribute properly also without using ME (column 5). In fact, using the single layer RN with ME decreases its performance (column 6). For architectures with multiple relational layers, ME is still crucial.

With 3 layers, ME and L2 regularization we obtain a total accuracy of 96.41\% (column 4). Only XOR is not yet above 90\%. A significant final step yields the LAMB optimizer. The total accuracy with 3 layers, ME, L2 regularization and LAMB is 98.03\% (last column). Every single accuracy now exceeds 95\%, except for XOR with 93.89\%, but most of them by far. 

\begin{table*}
    \caption{Performance comparison with and without Magnitude Encoding (ME). Models in the left section are trained using Adam, models on the right use the LAMB optimizer. Vanilla WReN results use dropout instead of L2 regularization. The three sections show the test accuracies for single categories.}
    \centering
    \begin{tabularx}{\linewidth}{c|L L L L|L L L L L L}
        Architecture & vanilla \mbox{WReN \cite{barrett18a}} & 1-layer MLRN & 3-layer MLRN & 3-layer MLRN & 1-layer MLRN & 1-layer MLRN & 2-layer MLRN & 2-layer MLRN & 3-layer MLRN & 3-layer MLRN \\
        Using L2 reg. & $\times$ & \checkmark & \checkmark & \checkmark & \checkmark & \checkmark & \checkmark & \checkmark & \checkmark & \checkmark \\
        Using ME & $\times$ & $\times$ & $\times$ & \checkmark & $\times$ & \checkmark & $\times$ & \checkmark & $\times$ & \checkmark \\
        Using LAMB & $\times$ & $\times$ & $\times$ & $\times$ & \checkmark & \checkmark & \checkmark & \checkmark & \checkmark & \checkmark \\
        \hline
        \hline
line & 78.3 & 97.08 & 97.96 & \textbf{99.26} & 98.18 & 97.61 & 99.20 & 99.00 & 99.09 & 98.84 \\
shape & 46.2 & 71.33 & 73.92 & 93.30 & 89.83 & 86.37 & 80.96 & 94.92 & 81.30 & \textbf{96.86} \\
\hline
color & 58.9 & 71.95 & 71.83 & 93.82 & 91.79 & 86.79 & 73.58 & 93.72 & 73.60 & \textbf{96.71} \\
position & 77.3 & 93.56 & 99.26 & 99.57 & 97.50 & 96.81 & 99.68 & 99.57 & 99.70 & \textbf{99.76} \\
type & 61.0 & 91.24 & 92.67 & 97.08 & 94.73 & 95.42 & 99.51 & 99.27 & \textbf{99.67} & 98.41 \\
number & 80.1 & 99.08 & \textbf{99.86} & 99.59 & 99.04 & 97.41 & 99.77 & 99.31 & 98.63 & 99.85 \\
size & 26.4 & 77.66 & 83.86 & 95.33 & 90.56 & 87.63 & 94.91 & 95.10 & \textbf{96.56} & 95.91 \\
\hline
AND & 63.2 & 86.20 & 86.68 & 96.27 & 95.91 & 93.05 & 90.26 & 96.58 & 90.43 & \textbf{97.52} \\
cons\_union & 60.1 & 91.09 & 90.75 & 99.58 & 98.66 & 98.10 & 91.99 & 99.37 & 91.52 & \textbf{99.68} \\
XOR & 53.2 & 69.22 & 78.05 & 89.60 & 81.50 & 78.29 & 88.16 & 91.76 & 89.02 & \textbf{93.89} \\
OR & 64.7 & 86.85 & 86.94 & 97.02 & 95.84 & 92.94 & 90.95 & 98.09 & 91.27 & \textbf{98.74} \\
progression & 55.4 & 87.89 & 86.99 & 99.36 & 98.93 & 98.60 & 88.38 & 99.36 & 88.00 & \textbf{99.69} \\
\hline
\hline
All single acc & 68.5 & 84.17 & 85.90 & 96.27 & 94.00 & 91.97 & 90.05 & 96.95 & 90.17 & \textbf{97.85} \\
\hline
Total acc & 62.6 & 82.31 & 84.28 & 96.41 & 94.14 & 92.15 & 88.53 & 97.18 & 88.60 & \textbf{98.03} \\
\hline
\hline
Total error & 37.4 & 17.69 & 15.72 & 3.59 & 5.86 & 7.85 & 11.47 & 2.82 & 11.40 & \textbf{1.97} \\
    \end{tabularx}
    \label{tab:finalres}
\end{table*}

\subsection{Local Minima}
\begin{figure*}
    \centering
    \begin{tikzpicture}
        \begin{axis}[
            xlabel=$epoch$,
            ylabel=$accuracy$,
            ymin=0.8,
            grid=both,
            max space between ticks=34pt,
            minor grid style={gray!25},
            major grid style={gray!25},
            width=0.9\textwidth,
            height=0.49\textwidth,
            no marks,
            legend style={at={(0.95,0.05)}, anchor=south east}, 
            legend columns=2]
        \addplot[line width=1pt,dashed,color=red] %
            table[x=epoch,y=training_acc,col sep=semicolon]{events.out.tfevents.1565626674.pc47.training.csv};
        \addlegendentry{Training run 1};
        \addplot[line width=1pt,solid,color=red] %
            table[x=epoch,y=validation_acc,col sep=semicolon]{events.out.tfevents.1565626674.pc47.validation.csv};
        \addlegendentry{Validation run 1};
        \addplot[line width=1pt,dashed,color=orange] %
            table[x=epoch,y=training_acc,col sep=semicolon]{events.out.tfevents.1565781975.pc47.training.csv};
        \addlegendentry{Training run 2};
        \addplot[line width=1pt,solid,color=orange] %
            table[x=epoch,y=validation_acc,col sep=semicolon]{events.out.tfevents.1565781975.pc47.validation.csv};
        \addlegendentry{Validation run 2};
        \addplot[line width=1pt,dashed,color=green] %
            table[x=epoch,y=training_acc,col sep=semicolon]{events.out.tfevents.1565885707.pc47.training.csv};
        \addlegendentry{Training run 3};
        \addplot[line width=1pt,solid,color=green] %
            table[x=epoch,y=validation_acc,col sep=semicolon]{events.out.tfevents.1565885707.pc47.validation.csv};
        \addlegendentry{Validation run 3};
        \addplot[line width=1pt,dashed,color=blue] %
            table[x=epoch,y=training_acc,col sep=semicolon]{events.out.tfevents.1569318989.pc47.training.csv};
        \addlegendentry{Training run 4};
        \addplot[line width=1pt,solid,color=blue] %
            table[x=epoch,y=validation_acc,col sep=semicolon]{events.out.tfevents.1569318989.pc47.validation.csv};
        \addlegendentry{Validation run 4};
        \addplot[line width=1pt,dashed,color=teal] %
            table[x=epoch,y=training_acc,col sep=semicolon]{events.out.tfevents.1569436488.pc47.training.csv};
        \addlegendentry{Training run 5};
        \addplot[line width=1pt,solid,color=teal] %
            table[x=epoch,y=validation_acc,col sep=semicolon]{events.out.tfevents.1569436488.pc47.validation.csv};
        \addlegendentry{Validation run 5};
        \end{axis}
    \end{tikzpicture}
    \caption{Training and validation curves for 3-layer MLRN based models with L2 regularization, ME20 and LAMB optimizer. All runs used the same hyperparameters.  Nevertheless, the same model can converge towards two distinct states with distinct performance levels. The lower performance level is indistinguishable from non-ME models.}
    \label{fig:traincurve}
\end{figure*}
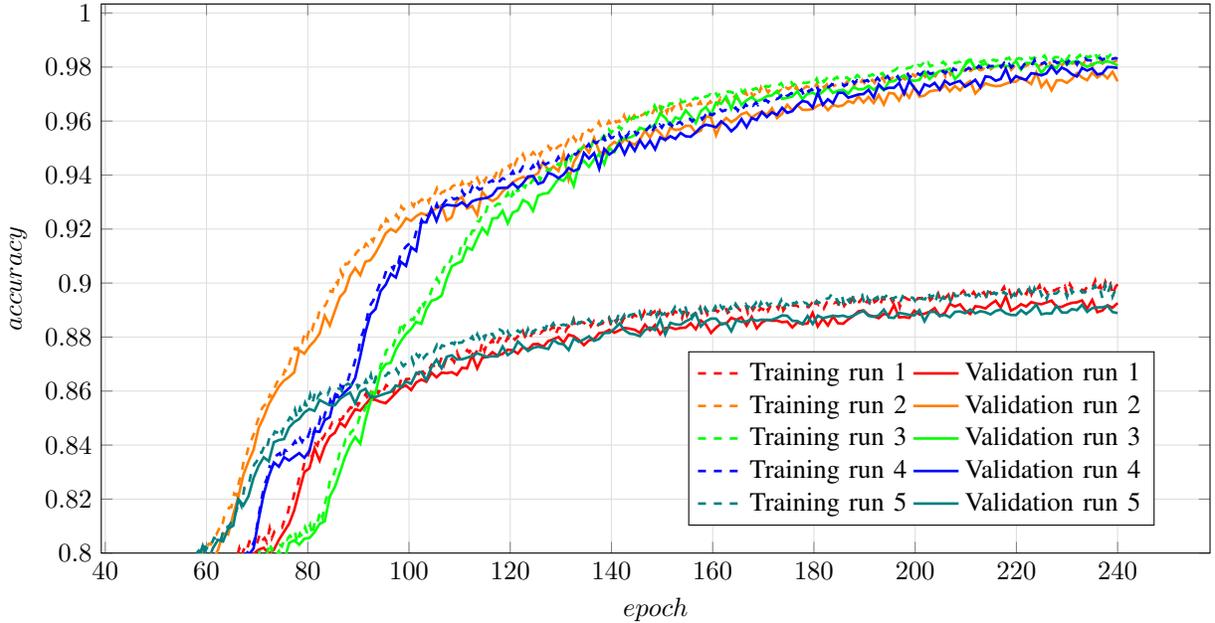

When training models with ME, two distinct performance levels can be observed, each with a probability of about 50\%. This is the case for both, Adam and LAMB optimizer. Figure \ref{fig:traincurve} shows training and validation curves for different runs. The two points of convergence are not only different in test performance, but the training curves also exhibit the two distinct levels. This means there are two points in the parameter space that the model can converge to, so the two points behave like local minima. 
One of them is indistinguishable from the non-ME model, showing poor performance on the $color$ attribute, while the other displays the expected improvement in performance. 
This phenomenon is likely owed to the fact that the scalar representation can easily be reconstructed, or at least approximated, from its vectorial representation by a linear layer.

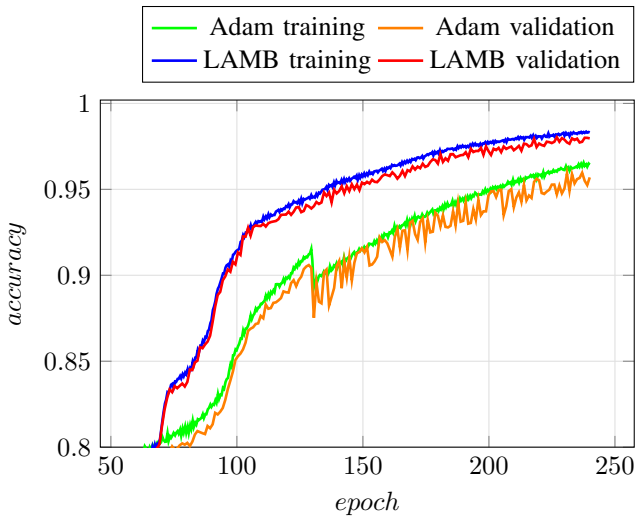
\begin{figure}
    \centering
    \begin{tikzpicture}
        \begin{axis}[
            xlabel=$epoch$,
            ylabel=$accuracy$,
            ymin=0.8,
            grid=both,
            minor grid style={gray!25},
            major grid style={gray!25},
            width=0.98\linewidth,
            height=0.7\linewidth,
            no marks,
            legend style={at={(1.0,1.05)}, anchor=south east}, 
            legend columns=2]
        \addplot[line width=1pt,solid,color=green] %
            table[x=epoch,y=training_acc,col sep=semicolon]{events.out.tfevents.1561908914.pc47.training.csv};
        \addlegendentry{Adam training};
        \addplot[line width=1pt,solid,color=orange] %
            table[x=epoch,y=validation_acc,col sep=semicolon]{events.out.tfevents.1561908914.pc47.validation.csv};
        \addlegendentry{Adam validation};
        \addplot[line width=1pt,solid,color=blue] %
            table[x=epoch,y=training_acc,col sep=semicolon]{events.out.tfevents.1569318989.pc47.training.csv};
        \addlegendentry{LAMB training};
        \addplot[line width=1pt,solid,color=red] %
            table[x=epoch,y=validation_acc,col sep=semicolon]{events.out.tfevents.1569318989.pc47.validation.csv};
        \addlegendentry{LAMB validation};
        \end{axis}
    \end{tikzpicture}
    \caption{Training and validation curves for 3-layer MLRN based models with L2 regularization and ME20. Training with the Adam optimizer exhibits significantly more noise in the validation curve compared to training with the LAMB optimizer. The kink in the curves for the Adam optimizer occurs on every run.}
    \label{fig:optimcurve}
\end{figure}

\subsection{Remarks on Generalization}
The results discussed so far only involved test samples drawn from the same data distribution as the training set. This is the so-called "neutral split".
However, since cognitive abilities are often associated with learning first principles and transfering them to other problem domains, the PGM dataset also contains training sets with some problem classes withheld to test more advanced generalization capabilities. 

\begin{table}
    \caption{Comparison of generalization performance in three different regimes.}
    \centering
    \begin{tabular}{c|c c c}
        Generalization & \makecell{neutral} & \makecell{interpolation} & \makecell{extrapolation} \\
        \hline
        \makecell{vanilla WReN \cite{barrett18a}}  & 62.6 & 64.4 &17.2 \\
        \makecell{3-layer  MLRN+L2 +ME}  & 98.0 & 57.8 & 14.9 \\
    \end{tabular}
    \label{tab:generalization}
\end{table}

Table \ref{tab:generalization} shows that the new model optimized for the neutral generalization regime is even slightly worse for the more advanced generalization tests than the original WReN. 
It remains for future work to investigate whether the model changes discussed in this paper are sufficient to achieve good results for the other regimes, if hyperparameter search is done on their respective training and validation sets. 
Also, the fact that the models with more relational layers are not showing much improvement in performance might hint at problems lying in overfitting of the embedding layers.

\section{Discussion}

So far several benchmark datasets for relational reasoning have been introduced. Most of them could be solved by purely neural network based models, except for the PGM dataset based on Raven's Progressive Matrices. With this work that benchmark is also solved now, at least for the standard setting, the so-called "neutral split". Compared to the state-of-the-art, the error could be reduced by a factor of 20. This is achieved by combining alternative regularization, a new optimizer, additional relational layers and a vectorial encoding of scalar inputs (pixel colors). 

The PGM dataset has an additional emphasis on learning the transfer of first principles by testing the performance on relations or attributes that are withheld from the training set. This is still an open problem and has also not yet been solved with our approach. Nevertheless, while this remains a long-term goal, examining models which perform well on the neutral test set is a prerequisite to determine how shortcomings in learning first principles can be overcome. 

The experiments show a vast difference in difficulty between various relation types. The results are further evidence that Multi-Layer Relation Networks are capable of learning more complex relations. It is further shown that late fusion models can have difficulties preserving low level features up to the point where the features are fused. 
This is mitigated by using a data agnostic encoding scheme, but the local minima that are introduced might make early fusion models more desirable instead.

Lastly, using the LAMB optimizer rather than Adam not only improves the test accuracy by a significant margin, but also makes the training loss and accuracy curves smoother, as can be seen in Fig. \ref{fig:optimcurve}. Whether this can be attributed to the vastly different activation levels in the relation layers remains to be studied in more detail.

In combination the techniques solve the PGM benchmark on the neutral generalization regime beyond 98.0\% test accuracy, which makes it reasonable to shift the focus to the other generalization regimes going forward. In the meantime, a second benchmark dataset based on Raven's Progressive Matrices was introduced, called the RAVEN dataset \cite{zhang19}, which seems to be a good candidate to test even harder generalization problems in abstract reasoning without using different distributions for training and test data. It should be noted that while these kinds of benchmarks are not exactly comparable to human IQ tests due to the large number of training samples, they are still an important measure for the potency of models on abstract reasoning tasks.

\bibliography{paperbib.bib}{}
\bibliographystyle{plain}
\vspace{12pt}

\end{document}